\documentclass[11pt]{article}%
\usepackage{amsmath}%
\usepackage{amsfonts}%
\usepackage{amssymb}%
\usepackage{graphicx}

\usepackage{array}

\usepackage{multirow}
\usepackage{color}
\usepackage{colortbl}
\usepackage[table]{xcolor}
\usepackage{hyperref}
\usepackage[margin=1in]{geometry}
\usepackage{tabularx, booktabs}

\definecolor{LightCyan}{rgb}{0.88,1,1}
\definecolor{Gray}{gray}{0.85}

\newcolumntype{Y}{>{\centering\arraybackslash}X}

\usepackage{array}
\newcolumntype{L}[1]{>{\raggedright\let\newline\\\arraybackslash\hspace{0pt}}m{#1}}
\newcolumntype{C}[1]{>{\centering\let\newline\\\arraybackslash\hspace{0pt}}m{#1}}
\newcolumntype{R}[1]{>{\raggedleft\let\newline\\\arraybackslash\hspace{0pt}}m{#1}}

\begin{document}

\title{Process Discovery using Inductive Miner and Decomposition.\\A Submission to the Process Discovery Contest @ BPM2016.\\Technical Report}
\author{Raji Ghawi
\\American University of Beirut, Lebanon}
\date{October 2016}
\maketitle

\begin{abstract}
This report presents a submission to the Process Discovery Contest \cite{contest}. The contest is dedicated to the assessment of tools and techniques that discover business process models from event logs. The objective is to compare the efficiency of techniques to discover process models that provide a proper balance between ``overfitting" and ``underfitting".
In the context of the Process Discovery Contest, process discovery is turned into a classification task with a training set and a test set; where a process model needs to decide whether traces are fitting or not.

In this report, we first show how we use two discovery techniques, namely: \emph{Inductive Miner} and \emph{Decomposition}, to discover process models from the training set using ProM tool. Second, we show how we use replay results to 1) check the rediscoverability of models, and to 2)  classify unseen traces (in test logs) as fitting or not. Then, we discuss the classification results of validation logs, and the complexity of discovered models, and their impact on the selection of models for submission. The report ends with the pictures of the submitted process models.
\end{abstract}

\section{Introduction}

Process Mining is a recently emerging discipline that links data analysis with process management. One of the central tasks of Process Mining is \textbf{Process Discovery}, where knowledge is extracted from event logs (readily available in today's information systems) in order to discover real processes \cite{pm}. Several tools, techniques, and algorithms have been proposed to discover business process models from event logs. The quality of discovered models and the efficiency of discovery techniques vary according to several criteria including: fitness, precision, generalization and simplicity.

This document is our submission to the Process Discovery Contest organized by the IEEE CIS Task Force on Process Mining\footnote{\url{http://www.win.tue.nl/ieeetfpm/doku.php}}, and which is collocated with the BPM Conference in Rio de Janeiro in September 2016.

The contest is dedicated to the assessment of tools and techniques that discover business process models from event logs. The objective is to compare the efficiency of techniques to discover process models that provide a proper balance between ``overfitting" and ``underfitting". A process model is overfitting (the event log) if it is too restrictive, disallowing behavior which is part of the underlying process. 
Conversely, it is underfitting (the reality) if it is not restrictive enough, allowing behavior which is not part of the underlying process. 
 
Within the Process-Discovery contest, 10 event logs are provided to the contestants as a training set. These event logs are generated from business process models that show different behavioral characteristics. The process models are not disclosed to the contestants.
The contestants should use the training event logs to discover process models that are the closest to the original process models, in term of balancing between ``overfitting" and ``underfitting".
To assess this balance, the contest has a classification perspective where a test event log will be used.
The test event log contains traces representing real process behavior and traces representing behavior not related to the process. 
A model is as good in balancing ``overfitting" and ``underfitting" as it is able to correctly classify the traces in the ``test" event log:
\begin{itemize}
	\item Given a trace representing real process behavior, the model should classify it as allowed.
	\item Given a trace representing a behavior not related to the process, the model should classify it as disallowed.
\end{itemize}

With a classification view, the winner is the contestant who can classify correct the largest number of traces in all the test event logs. Moreover, in order to help the contestants tune their discovered models, two ``validation" sets are made available in April and May 2016.

In our submission to the contest, we discovered process models using two techniques: Inductive Miner \cite{indc} and Decomposition based discovery \cite{decmp1, decmp2, decmp3}. We chose these techniques because they guarantee rediscoverability; that is, models discovered using those techniques \emph{fit} (generate) all the traces in their original training event logs.
We used ProM tool (\url{http://www.promtools.org/}) to apply those techniques in discovering our submitted models.

In the reminder of this report, we first show in Section \ref{sec:2} how do we use two discovery techniques, namely: \emph{Inductive Miner} and \emph{Decomposition}, to discover process models from the training set using ProM tool. Second, we present in Section \ref{sec:3} how we \emph{replay} an event log on a process model. Then, we show how we use replay results:
\begin{enumerate}
	\item to check the rediscoverability of models (Section \ref{sec:31}), and 
	\item to classify unseen traces (in validation and/or test logs) as fitting or not (Section \ref{sec:class}).
\end{enumerate}
Section \ref{sec:discussion} is dedicated for a discussion about the classification results of April and May validation logs, the complexity of discovered models, and how these two points affect our selection of models for submission.
In the end of the report, we show the pictures of our submitted models.


\section{Process Discovery}
\label{sec:2}

We used two well known discovery techniques: Inductive Miner \cite{indc} and Decomposed Process Mining \cite{decmp1, decmp2, decmp3}. We used those techniques to discover process models for each one of the 10 training event logs. However, since only one process model should be submitted for every event log, we decided to include in our submission a mix of those models.

Rationally, we have chosen those techniques because they guarantee a level of \emph{\textbf{rediscoverability}}, namely the ability of discovering again the models used to generate the traces of the event log. We consider that \textbf{rediscoverability} property is essential to accomplish the task of correctly classify new unseen traces.
 
The models for processes 1, 2, 4, 8 and 9 are discovered using \textbf{Inductive Miner}. We used ProM Lite \footnote{\url{http://www.promtools.org/doku.php?id=promlite}} for this group of models. Other models - for processes 3, 5, 6, 7, and 10 - are discovered using \textbf{Decomposition} technique. We used ProM 6.5.1 \footnote{\url{http://www.promtools.org/doku.php?id=prom651}} for this group of models. The justification of this selection of process models is given in Section \ref{sec:discussion}. 

In the reminder of this section, we describe the step-by-step procedure to discover process models using the above mentioned techniques.


\subsection{Discovery using Inductive Miner}
Inductive Miner \cite{indc} is a discovery approach to construct a Process Tree for a given log.
The main advantages of this approach are:
\begin{itemize}
	\item All discovered models correspond to sound, block-structured workflow (WF) net systems.
	\item The model always fits, i.e., the model can generate the traces in the log.
\end{itemize}
The approach works recursively with \emph{divide and conquer} strategy: split log, construct part of process tree, and proceed with handling split parts of log separately. A process tree can be directly transformed into a Petri Net.
However, ProM has a plugin that can mine a Petri net directly using Inductive Miner technique.



To discover a model (Petri net) from an event log with Inductive Miner technique, we used ProM Lite\footnote{The Inductive Miner plugin is also available in ProM 6.5.1. However, we get different models when we used Inductive Miner in ProM Lite and ProM 6.5.1 (apparently, due to implementation discrepancies). Therefore, we would emphasize that we used \textbf{ProM Lite} for the submitted models that are discovered using Inductive Miner (as mentioned above: models 1, 2, 4, 8, and 9).} and followed the following steps: 

\begin{enumerate}
	\item First, the log (in \textbf{XES} format) is loaded onto ProM and used.
Then, in \emph{Actions} window, the plugin ``\textbf{Mine Petri net with Inductive Miner}" by Leemans is selected (see Figure \ref{fig:discv_induc_1}).

\begin{figure}[htbp]
	\centering
		\includegraphics[width=0.5\textwidth]{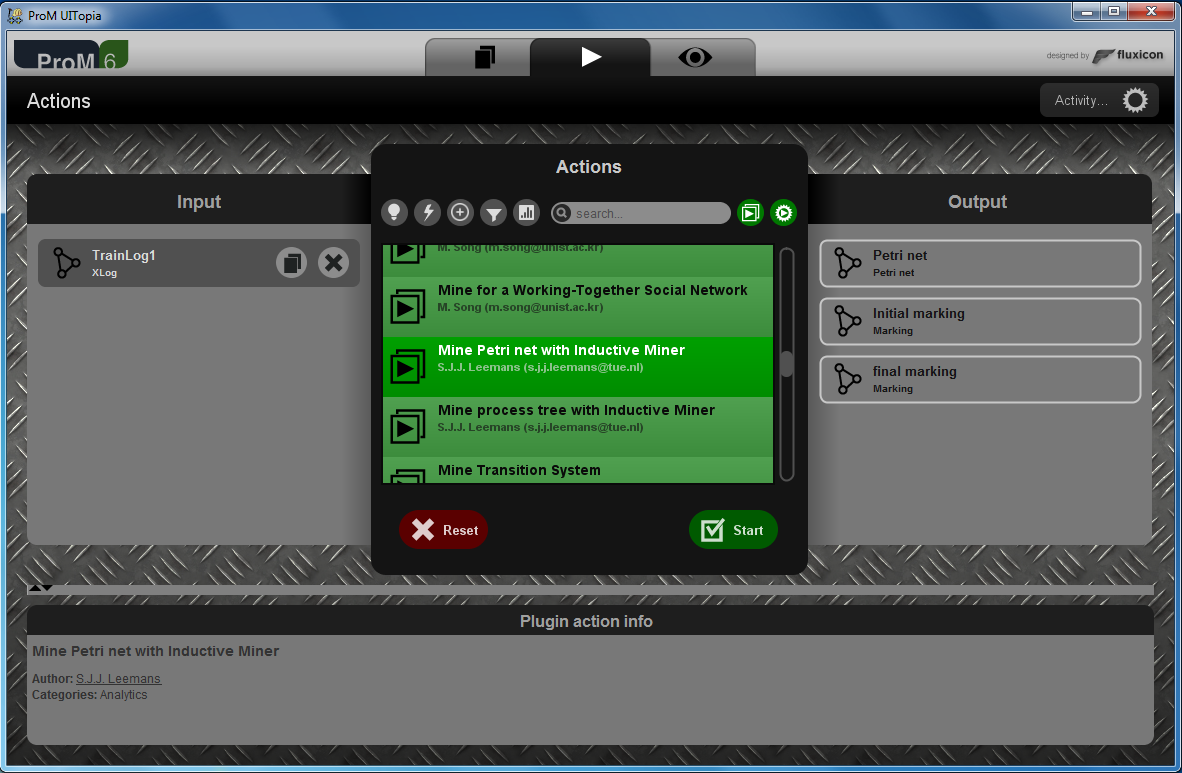}
	\caption{Discovery using Inductive Miner: Plugin Selection}
	\label{fig:discv_induc_1}
\end{figure}

	\item A settings page appears then (Figure \ref{fig:discv_induc_2}). For \textbf{Variant}, we select \emph{Inductive Miner}. For \textbf{Event Classifier}, we select \emph{Event Name}. 

\begin{figure}[htbp]
	\centering
		\includegraphics[width=0.5\textwidth]{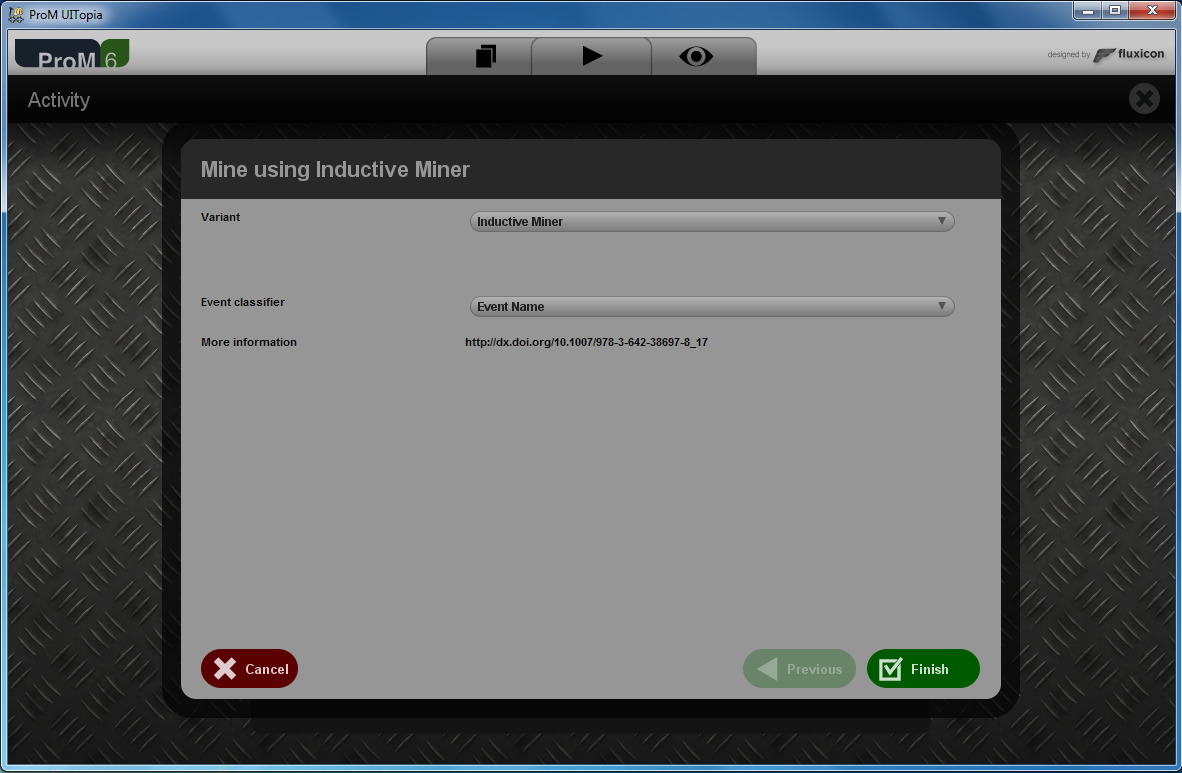}
	\caption{Discovery using Inductive Miner: Settings Selection}
	\label{fig:discv_induc_2}
\end{figure}

\item Finally, we get a Petri net discovered using Inductive Miner. The discovered model can be visualized using \emph{Graphviz Petri net visualization} (Figure \ref{fig:discv_induc_3}).

\begin{figure}[htbp]
	\centering
		\includegraphics[width=0.5\textwidth]{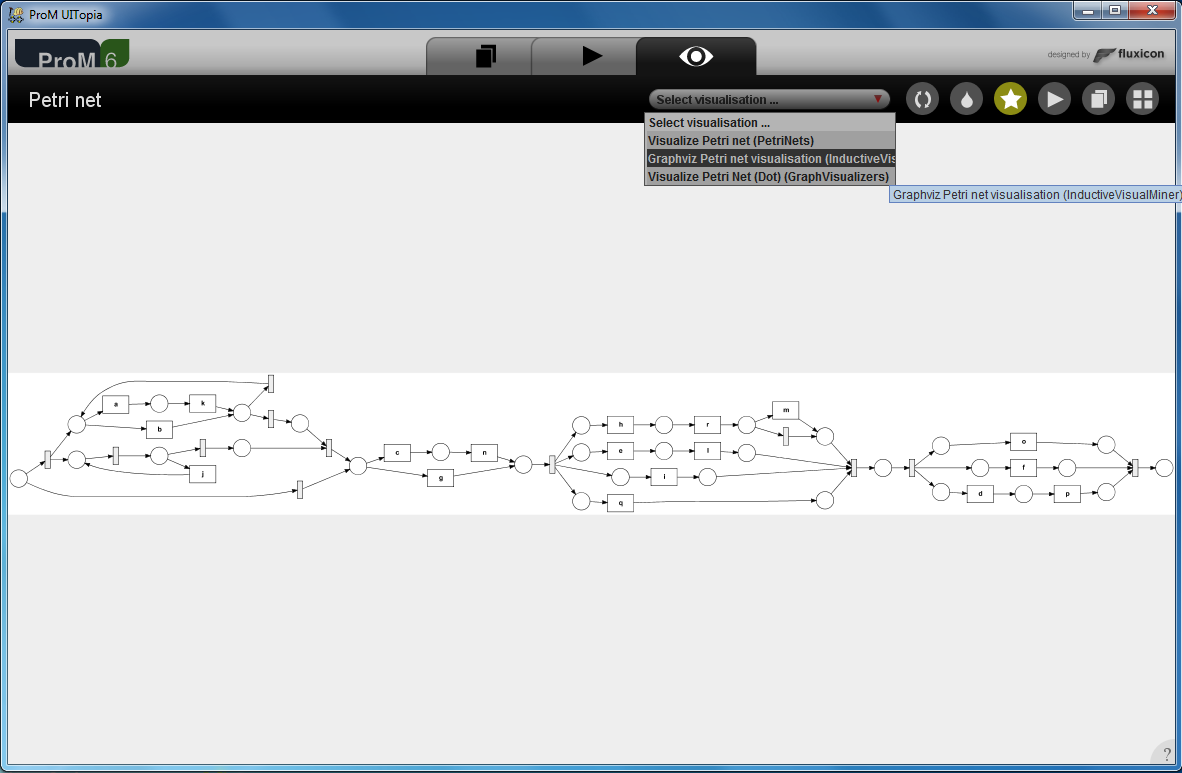}
	\caption{Discovery using Inductive Miner: Model Visualization}
	\label{fig:discv_induc_3}
\end{figure}

\end{enumerate}


\subsection{Discovery using Decomposition}
Decomposition \cite{decmp1, decmp2, decmp3} is a process mining technique that discovers an accepting Petri net\footnote{Accepting Petri net is a labeled Petri net with an initial marking and a collection of final markings} from a log using decomposition.
First, the log will be split into a number of sublogs. Second, a subnet will be discovered (using the given discovery algorithm) for every sublog.
Third, these subnets will be merged into a single accepting Petri net. Fourth, known reduction techniques will be applied on this accepting Petri net, and the result is returned.


To discover a model from an event log using Decomposition technique, we used ProM 6.5.1 
(this technique is not available in ProM Lite). 

\begin{enumerate}
	\item First, the log (in \textbf{XES} format) is loaded onto ProM and used.
Then, in \emph{Actions} window, the plugin ``\textbf{Discovery using Decomposition}" is selected (see Figure \ref{fig:discv_decomp_1}). Actually, there are two available plugins in ProM 6.5.1 for discovery using Decomposition: an \textbf{interactive} plugin that allows for custom settings, and a \textbf{batch} one that uses some \emph{default} settings. We used the \textbf{batch} plugin, hence default settings are used; for instance, the underlying discovery algorithm is ILP Miner \cite{werf}\cite{decmp2}.

\begin{figure}[htbp]
	\centering
		\includegraphics[width=0.5\textwidth]{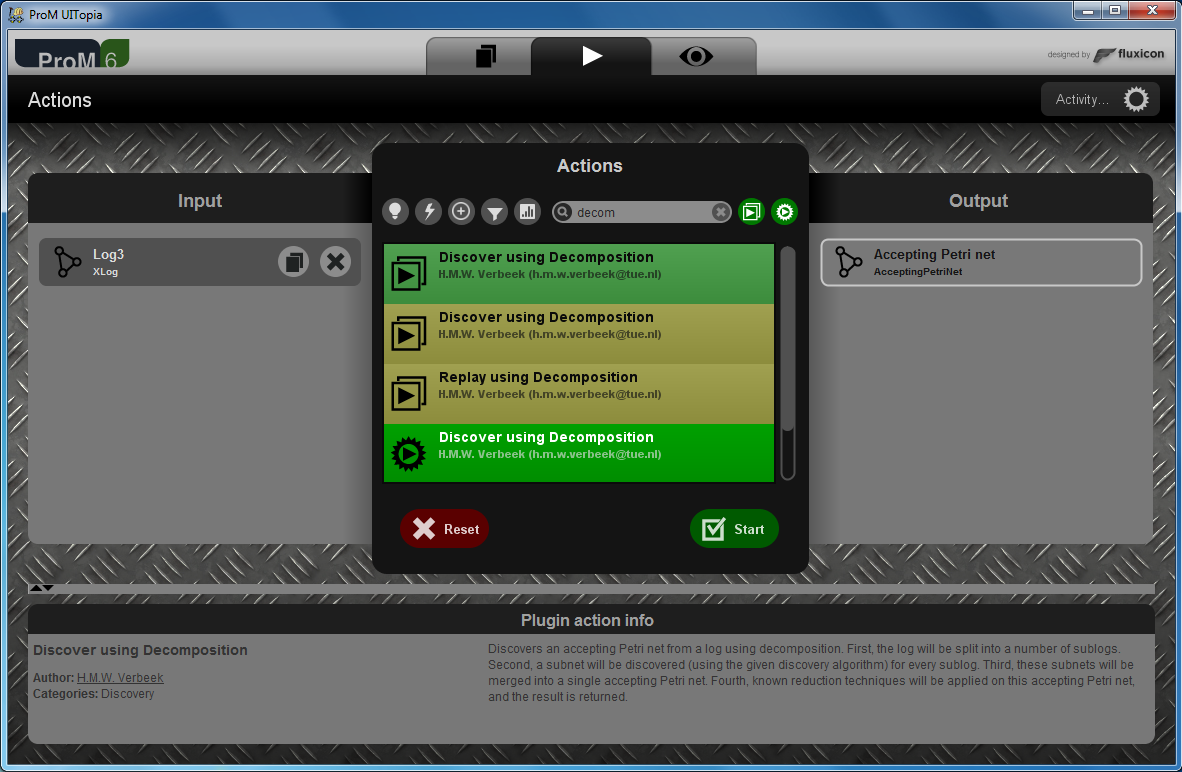}
	\caption{Discovery using Decomposition: Plugin Selection}
	\label{fig:discv_decomp_1}
\end{figure}

\item We get an accepting Petri net discovered using Decomposition. It can be visualized using different visualizations such as \emph{Visualize Accepting Petri Net (Dot)} (Figure \ref{fig:discv_decomp_2}). 

\begin{figure}[htbp]
	\centering
		\includegraphics[width=0.5\textwidth]{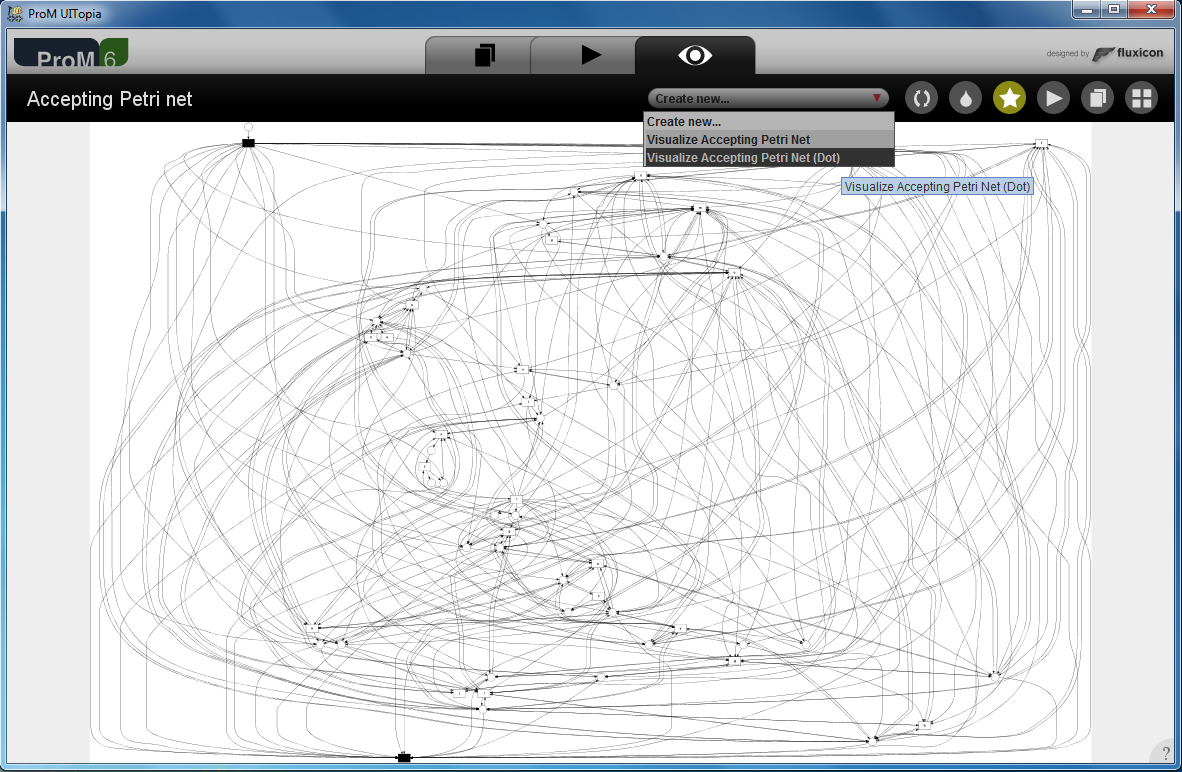}
	\caption{Discovery using Decomposition: Model Visualization}
	\label{fig:discv_decomp_2}
\end{figure}

\item Finally, we convert the discovered accepting Petri net into a `regular' Petri net using \textbf{Unpack Accepting Petri Net} plugin.

\end{enumerate}

Our models are all regular Petri nets that are exported as PNML files.


\section{Replaying Semantics}
\label{sec:3}

To replay an event log on a process model, we used the ProM plugin \textbf{Replay a Log on Petri Net for Conformance Analysis} (available in both ProM Lite and ProM 6.5.1, with identical results obtained).

We need \textbf{replay} for two tasks: 

\begin{enumerate}
	\item Verify that the discovered model can generate its originating event log (rediscoverability). 
	\item Classify new unseen traces (in validation and test datasets) as fitting or not-fitting with respect to the discovered models.
\end{enumerate}

To replay a log on a process model:

\begin{enumerate}
	\item First, in \emph{Workspace} page, both the log and the model (Petri net) are selected and used.
	\item Second, in \emph{Actions} page, the plugin \textbf{Replay a Log on Petri Net for Conformance Analysis} is selected (Figure \ref{fig:replay_1}).
		
\begin{figure}[ht]
	\centering
		\includegraphics[width=0.5\textwidth]{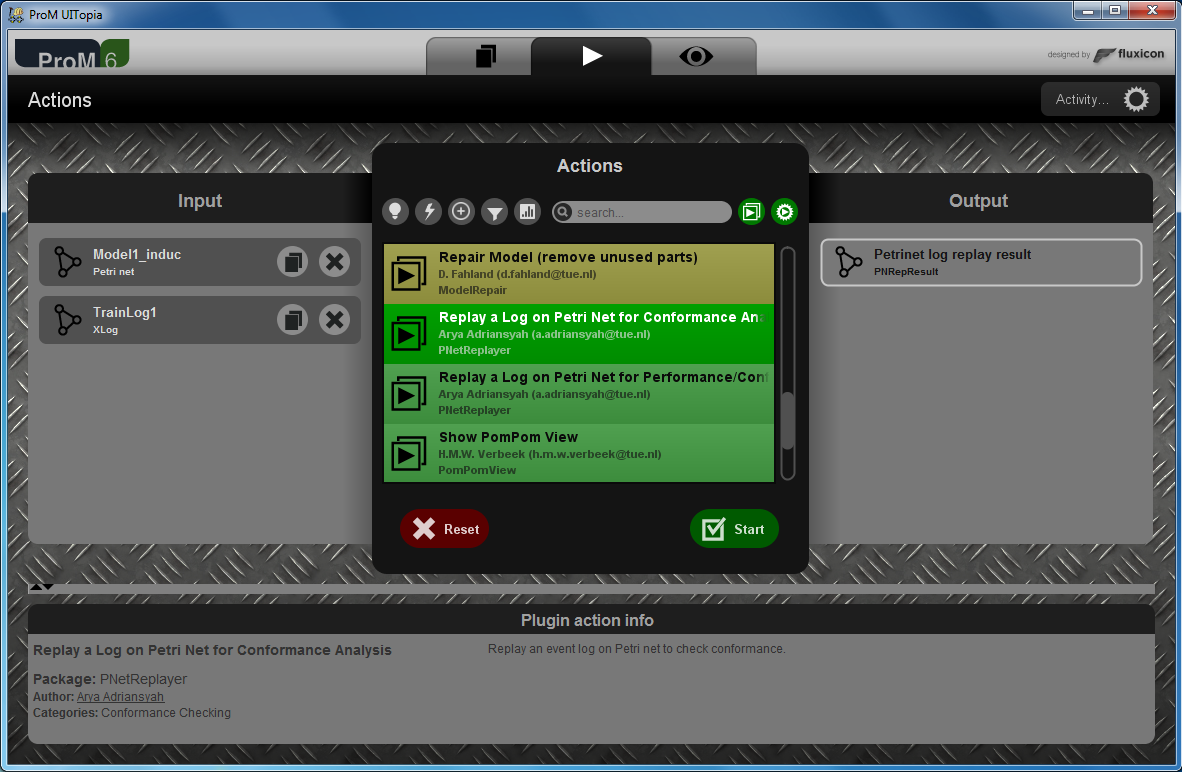}
	\caption{Replay Semantics: Plugin Selection}
	\label{fig:replay_1}
\end{figure}

\item Then, a dialog appears that allow for mapping between the transitions of the Petri net and the event classes of the log (Figure \ref{fig:replay_2}). We need only to choose \textbf{Event Name} as classifier; thus each transition is matched with the event class having the same name. 

\begin{figure}[ht]
	\centering
		\includegraphics[width=0.5\textwidth]{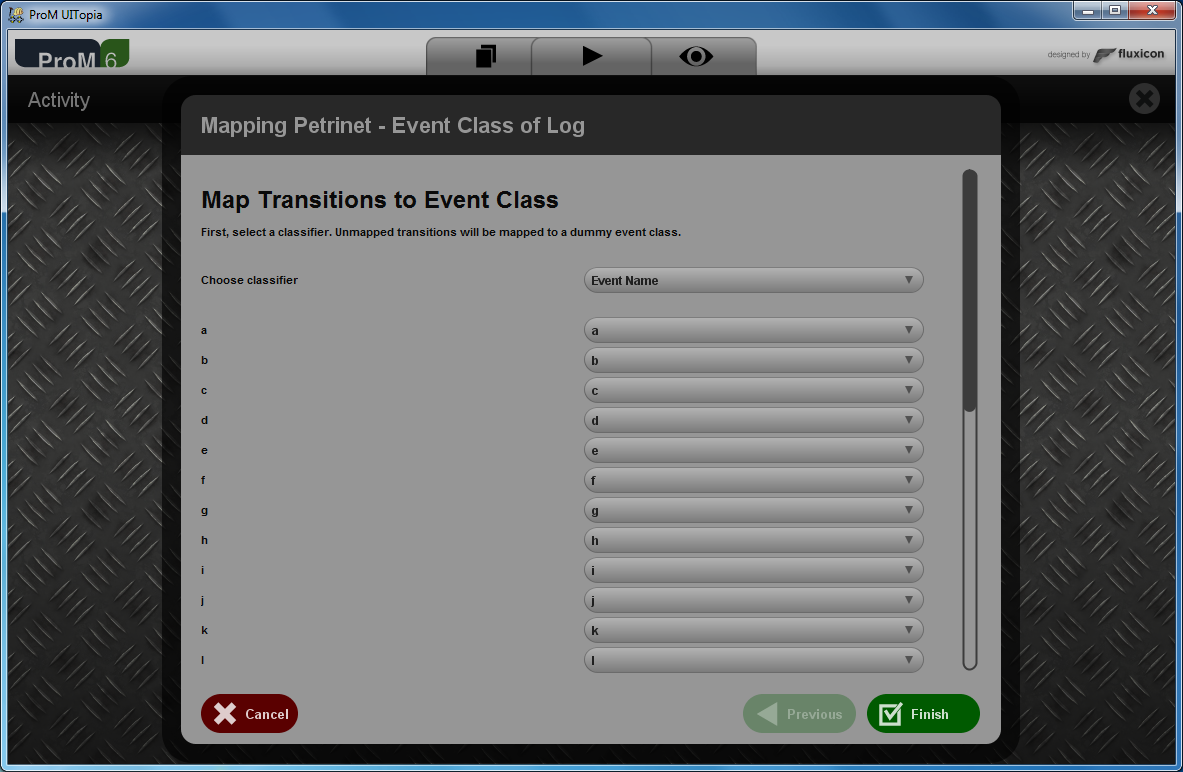}
	\caption{Replay Semantics: Mapping Petri Net to Log}
	\label{fig:replay_2}
\end{figure}



\item The next dialog allows us to select a replay algorithm. We use the following settings (as shown in Figure \ref{fig:replay_4}):
	
\begin{itemize}
	\item \textbf{The purpose of replay} -- measuring fitness.
	\item \textbf{Penalize improper completion}: yes.
	\item \textbf{Suggested algorithm} -- A* Cost-based Fitness Express, assuming at most 32767 tokens in each place.
\end{itemize}

\begin{figure}[ht]
	\centering
		\includegraphics[width=0.5\textwidth]{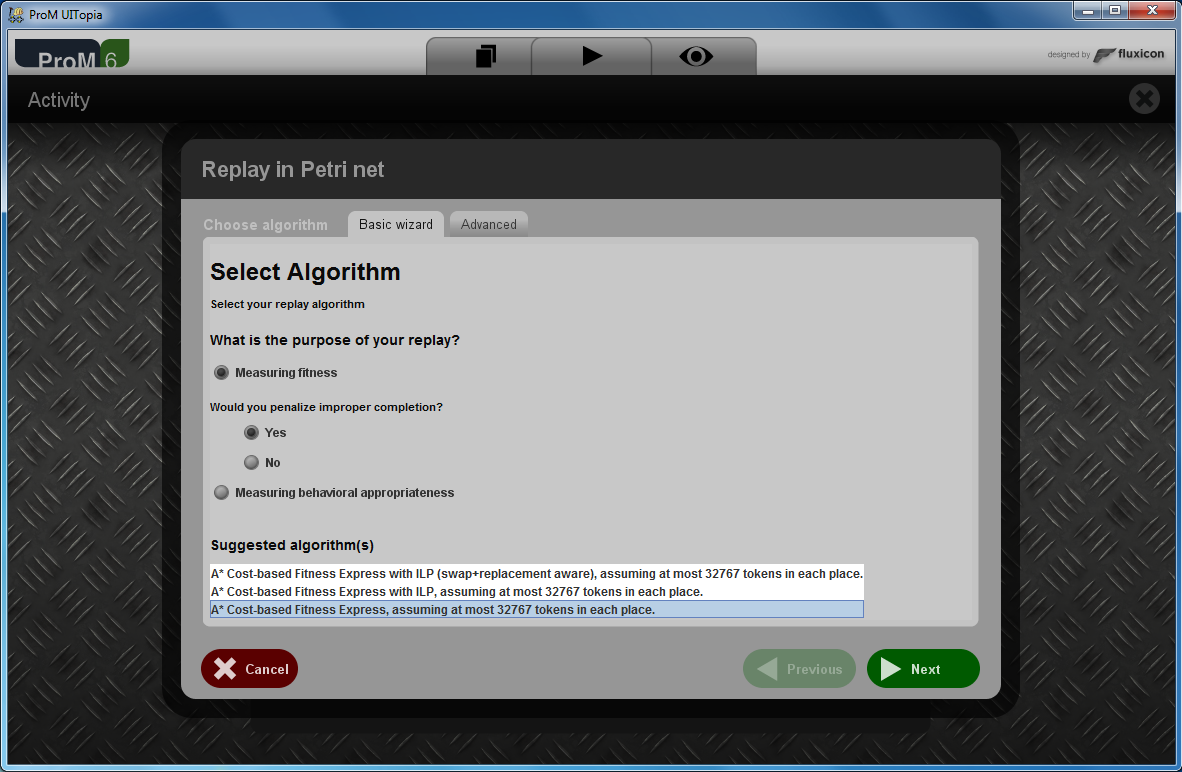}
	\caption{Replay Semantics: Algorithm Selection}
	\label{fig:replay_4}
\end{figure}

\item The last page of the dialog is for selecting the costs of particular deviation (Figure \ref{fig:replay_5}). Again, we keep the defaults: the cost is always 1 except for \emph{tau} transitions where the cost is 0.

\begin{figure}[ht]
	\centering
		\includegraphics[width=0.5\textwidth]{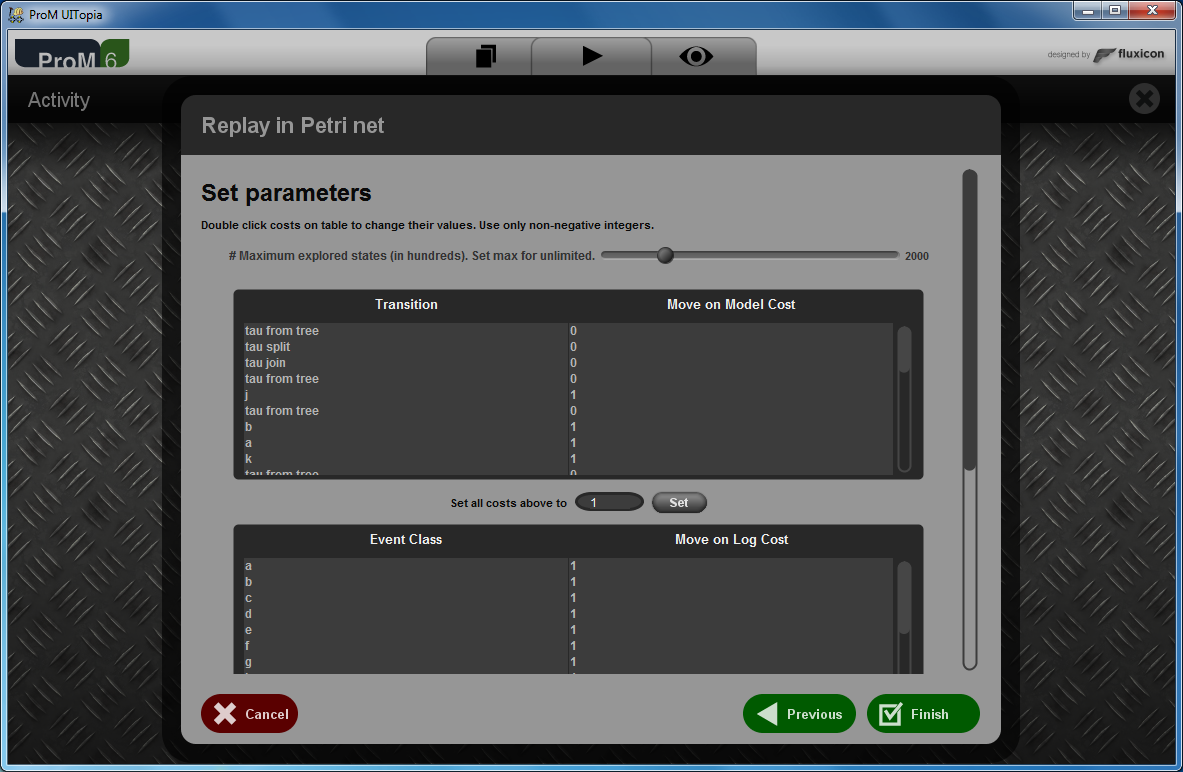}
	\caption{Replay Semantics: Deviation Cost Selection}
	\label{fig:replay_5}
\end{figure}

\end{enumerate}

Those are the required steps to replay a log on a process model. As mentioned above, we use those steps for rediscoverability checking and for classifying unseen traces.

\subsection{Rediscoverability Checking}
\label{sec:31}

Our choice of discovery techniques is mainly based on the rediscoverability property. That is, we chose Inductive Miner and Decomposition techniques because they both guarantee this property. Therefore, once we discover a model using one of those techniques, as discussed in Section \ref{sec:2}, we replay the originating/training event log on the discovered model to make sure that all the traces in the log \textbf{fit} properly in the model.

\begin{figure}[ht]
	\centering
		\includegraphics[width=0.5\textwidth]{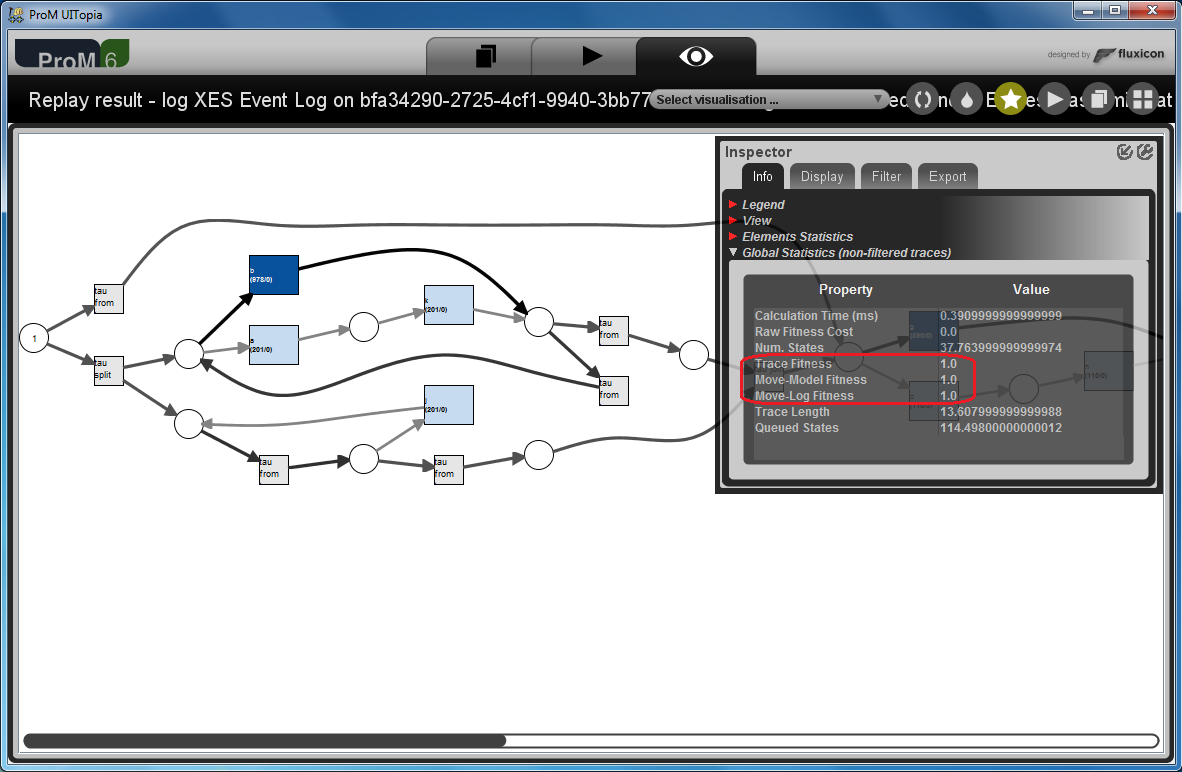}
	\caption{Replay Semantics: Rediscoverability Checking}
	\label{fig:replay_6}
\end{figure}

To do so, we follow the replay steps mentioned above using the training log and the discovered model. The result of the replay can be visualized in different ways.
For instance, using \textbf{Model Projected with Alignments} choice, the model can be visualized with alignments on top of it, as shown in Figure \ref{fig:replay_6}.

In this visualization, we can take a look at the \textbf{\emph{Global Statistics}} within the inspector window. Among the displayed statistics, we are mainly interested in three statistics: \textbf{Trace} Fitness, \textbf{Move-Model} Fitness, and \textbf{Move-Log} Fitness. The values of those statistics give an indicator about the rediscoverability property. 
When those global statistics are all equal to 1, this means that all the traces fit in the model; thus the model is able to rediscover the event log.
Otherwise, if anyone of those statistics has a value less than 1, then there are some traces in the log that don't fit in the model.

In our submission, all the models perfectly satisfy the rediscoverability property. 


\subsection{Classification of Unseen Traces}
\label{sec:class}

In the context of the Process Discovery Contest, process discovery is turned into a classification task with a training set and a test set. Moreover, two validation sets are made available to the contestants to allow them tune their discovered models. A process model needs to decide whether traces are fitting or not.

\begin{figure}[htb]
	\centering
		\includegraphics[width=0.76\textwidth]{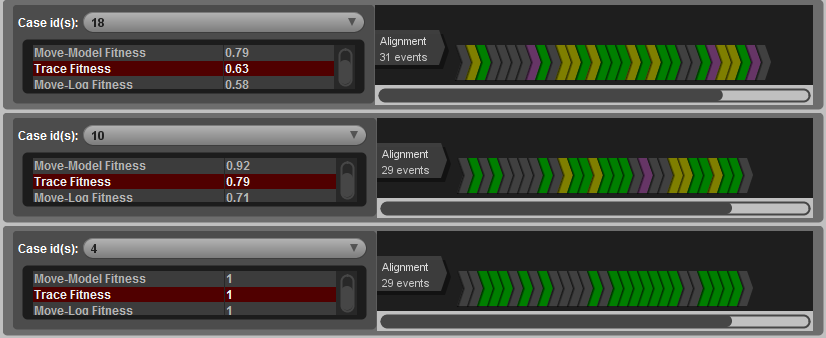}
	\caption{Replay results visualized using \textbf{Project Alignment to Log}}
	\label{fig:replay_results}
\end{figure}

For this classification task, a test or validation log is replayed on the discovered model.
Again, we used the ProM plugin \textbf{Replay a Log on Petri Net for Conformance Analysis} according to the steps mentioned earlier. 
When the replay results are obtained, we visualize them using \textbf{Project Alignment to Log} choice.
This will display log-model alignments for each case/trace, as shown in Figure \ref{fig:replay_results}.

\begin{figure}[htb]
	\centering
		\includegraphics[width=0.32\textwidth]{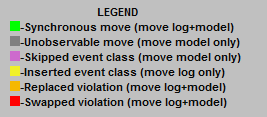}
	\caption{Alignment Legend}
	\label{fig:legend}
\end{figure}

For each trace, a table of statistics is shown to the left; and a colored picture of the trace alignment is shown to the right. The meaning of colors is illustrated in a legend shown on the right-top of the screen (Figure \ref{fig:legend}).

Among the displayed statistics, we are interested in: Trace fitness, Move-Log fitness, and Move-Model fitness; and we use them in the classification task.
We classify each trace as \emph{fitting} if, and only if, those three values are perfect (equal to 1). Otherwise, when anyone of those values is less than 1, then we classify the trace as \emph{not-fitting}.

For example, table \ref{tab:replay1} shows those values for replaying Log-1 of April validation dataset over our model-1 (discovered using Inductive Miner).

\begin{table}[htb]
	\centering
	 \scriptsize
		\begin{tabularx}{1\textwidth}{|l *{20}{|Y}|}	
\hline
Case ID    & 1 & 2 & 3 & 4 & 5 & 6 & 7 & 8 & 9 & 10 & 11 & 12 & 13 & 14 & 15 & 16 & 17 & 18 & 19 & 20 \\
\hline
Move-Model & 1 & 1 & .73 & 1 & .85 & .64 & .83 & 1 & .75 & .92 & 1 & .67 & .82 & 1 & .92 & 1 & 1 & .79 & 1 & 1 \\
Trace      & 1 & 1 & .73 & 1 & .87 & .67 & .83 & 1 & .74 & .79 & 1 & .65 & .82 & 1 & .88 & 1 & 1 & .63 & 1 & 1 \\
Move-Log   & 1 & 1 & .73 & 1 & .92 & .70 & .83 & 1 & .75 & .71 & 1 & .67 & .82 & 1 & .85 & 1 & 1 & .58 & 1 & 1 \\
\hline
Classification & $+$ & $+$ & $-$ & $+$ & $-$ & $-$ & $-$ & $+$ & $-$ & $-$ & $+$ & $-$ & $-$ & $+$ & $-$ & $+$ & $+$ & $-$ & $+$ & $+$ \\
\hline
	\end{tabularx}		
	\caption{Results of replaying Log-1 of April validation dataset over the model discovered using Inductive Miner; 
	Classification: $+$: fitting, $-$: non-fitting}
	\label{tab:replay1}
\end{table}

In the same sense, the colored picture of the trace alignment can be also used to determine fitness. If the picture consists of \textbf{green} and \textbf{gray} alignments \textbf{only}, then the trace is classified as \emph{fitting}. Otherwise, if the picture also contains other colors (like pink and/or yellow), then the trace is classified as \emph{not-fitting}.

\section{Discussion}
\label{sec:discussion}


In our solution to the Process Discovery Contest, we used Inductive Miner and Decomposition techniques - as presented in Section \ref{sec:2} - to discover process models\footnote{The pictures of process models of both groups are available at: \url{https://github.com/rajighawi/Process_Discovery_Contest_2016}} from the 10 \emph{training} logs. 
Both groups of models are then used to replay again the training logs - as presented in Section \ref{sec:31}- to ensure \emph{rediscoverability}. We found that all our process models \emph{satisfy the rediscoverability} property.

Then, we replayed April and May \emph{validation} event logs over both groups of process models - processes discovered using Inductive Miner and processes discovered using Decomposition. The replay results are then used to classify the traces of the validation logs as fitting or non-fitting - as presented in section \ref{sec:class}. We communicated the obtained classifications\footnote{The detailed classifications are available at: \url{https://github.com/rajighawi/Process_Discovery_Contest_2016}} with the contest organizers, who replied by indicating how many traces have been correctly classified for each one of the 10 logs. Table \ref{tab:april_results} shows the classification results of April validation logs using the two groups of process models: Inductive Miner and Decomposition. Analogue classification results of May logs are shown in Table \ref{tab:may_results}.

\begin{table}[h]
	\centering
	\small
		\begin{tabularx}{\textwidth}{ L{3.4cm}|YYYYYYYYYY|Y}
 Log $\rightarrow$& 1 & 2 & 3 & 4  & 5  & 6  & 7  & 8  & 9  & 10 &  \\
		\hline
		\multicolumn{12}{c}{Inductive Miner Process Models} \\		
		\hline
True Positive  & 10 & 10 & 10 & 10 & 10 & 10  & 10 & 10 & 10 & 10 & \\
False Positive & 0  & 0  & 8  & 0  & 1  & 3  & 0  & 5  & 0  & 0  & \\
True Negative  & 10 & 10 & 2  & 10 & 9  & 7  & 10 & 5  & 10 & 10 & \\
False Negative & 0  & 0  & 0  & 0  & 0  & 0  & 0  & 0  & 0  & 0  & \\
  \hline
Correctly Classified & 20 & 20 & 12 & 20 & 19 & 17 & 20 & 15 & 20 & 20 & \textbf{183}		\\	
		\hline
		\hline
		\multicolumn{12}{c}{Decomposition Process Models} \\		
		\hline
True Positive  & 10 & 10 & 10 & 10 & 10 & 10 & 10 & 10 & 10 & 10 & \\
False Positive & 0  & 0  & 1  & 0  & 0  & 1  & 0  & 6  & 0  & 0  & \\
True Negative  & 10 & 10 & 9  & 10 & 10 & 9  & 10 & 4  & 10 & 10 & \\
False Negative & 0  & 0  & 0  & 0  & 0  & 0  & 0  & 0  & 0  & 0  & \\
  \hline
Correctly Classified & 20 & 20 & 19 & 20 & 20 & 19 & 20 & 14 & 20 & 20 & \textbf{192}		\\	
		\end{tabularx}
	\caption{Classification results of \textbf{April} set using Inductive Miner and Decomposition.}
	\label{tab:april_results}
\end{table}


\begin{table}[h]
	\centering
	\small
		\begin{tabularx}{\textwidth}{ L{3.4cm}|YYYYYYYYYY|Y}
 Log $\rightarrow$& 1 & 2 & 3 & 4  & 5  & 6  & 7  & 8  & 9  & 10 &  \\
		\hline
		\multicolumn{12}{c}{Inductive Miner Process Models} \\		
		\hline
True Positive  & 10 & 10 & 10 & 10 & 10 & 10 & 10 & 10 & 10 & 10 & \\
False Positive & 0  & 0  & 9  & 0  & 1  & 4  & 3  & 2  & 0  & 2  & \\
True Negative  & 10 & 10 & 1  & 10 & 9  & 6  & 7  & 8  & 10 & 8  & \\
False Negative & 0  & 0  & 0  & 0  & 0  & 0  & 0  & 0  & 0  & 0  & \\
  \hline
Correctly Classified & 20 & 20 & 11 & 20 & 19 & 16 & 17 & 18 & 20 & 18 & \textbf{179}		\\	
		\hline
		\hline
		\multicolumn{12}{c}{Decomposition Process Models} \\		
		\hline
True Positive  & 10 & 10 & 10 & 10 & 10 & 10 & 10 & 10 & 10 & 10 & \\
False Positive & 0  & 0  & 0  & 0  & 1  & 0  & 1  & 4  & 0  & 0  & \\
True Negative  & 10 & 10 & 10 & 10 & 9  & 10 & 9  & 6  & 10 & 10 & \\
False Negative & 0  & 0  & 0  & 0  & 0  & 0  & 0  & 0  & 0  & 0  & \\
  \hline
Correctly Classified & 20 & 20 & 20 & 20 & 19 & 20 & 19 & 16 & 20 & 20 & \textbf{194}		\\	

		\end{tabularx}
	\caption{Classification results of \textbf{May} set using Inductive Miner and Decomposition.}
	\label{tab:may_results}
\end{table}


A quick look at these classification results tells that Decomposition models classify better than Inductive Miner models. That is, out of 200 traces in April logs, 192 are correctly classified using Decomposition models versus 183 using Inductive Miner models. Similarly, in May logs, 194 are correctly classified using Decomposition models versus 179 using Inductive Miner models

However, with a more careful examination of the classification results, we made the following observations:
\begin{itemize}
	\item Logs 1, 2, 4, and 9 are always classified correctly: 20 out of 20 traces are classified correctly.
	\item Log 8 has a better classification using Inductive Miner model than using Decomposition model (April: $15>14$, May: $18>16$).
	\item Other logs - namely 3, 5, 6, 7, and 10 - has a better classification using Decomposition models than using Inductive Miner models.
\end{itemize}

Now, based on these observations, we should decide which models shall we include in our submission to the contest\footnote{As mentioned earlier, only one model per log must be submitted.}. From the second and third observation, we can easily decide to consider Inductive Miner model for log: 8, and Decomposition models for logs: 3, 5, 6, 7 and 10.

However, the first observation tells us that neither one of the two groups of process models has a preference over the other with respect to the classification of logs: 1, 2, 4 and 9. Therefore, we have to consider another quality criteria of process models, namely: \emph{simplicity}.

Actually, the Inductive Miner models are generally more \emph{simple} than Decomposition ones. One can clearly see this by looking at the pictures of both groups of models (see Appendix \ref{apx:models}). However, in order to prove this claim we computed different complexity metrics of process models \cite{lassen, blum}. Table \ref{tab:metrics} shows the complexity metrics for both Inductive Miner and Decomposition process models. These metrics are: \emph{ECaM}: Extended Cardoso Metric, \emph{ECyM}: Extended Cyclomatic Metric, $|E|$ and $|V|$: number of Edges and Nodes in the reachability graph, $|A|$, $|P|$ and $|T|$: number of Arcs, Places, and Transitions in the Net, and \emph{SM}: Structuredness Metric.

Based on the values listed in Table \ref{tab:metrics} we can see that the Inductive Miner models are, in general, \emph{simpler} than Decomposition ones. For instance, the average structuredness of Decomposition models is roughly one order of magnitude higher than for Inductive Miner models.

\begin{table}[ht]
\centering
\small
\begin{tabular}{c|c|c|c|cc|ccc|c}			
Model & Density  
& \emph{ECaM}
& \emph{ECyM}
& $|E|$ & $|V|$ 
& $|A|$ & $|P|$ & $|T|$
& \emph{SM}  \\
\hline
\multicolumn{10}{c}{Inductive Miner Process Models} \\
\hline
1 & 0.0387 & 35 & 177  & 185   & 78   & 72 & 31 & 30 & 293   \\
2 & 0.0589 & 20 & 67   & 85    & 27   & 70 & 18 & 33 & 6180  \\
3 & 0.0795 & 18 & 51   & 65    & 20   & 62 & 13 & 30 & 1550  \\
4 & 0.0376 & 31 & 174  & 174   & 77   & 70 & 31 & 30 & 212   \\
5 & 0.0499 & 24 & 215  & 247   & 50   & 90 & 22 & 41 & 29205 \\
6 & 0.1038 & 12 & 23   & 31    & 10   & 54 & 10 & 26 & $ ^{**}$\\
7 & 0.0586 & 22 & 3730 & 4082  & 482  & 82 & 20 & 35 & 4160  \\
8 & 0.0536 & 24 & 159  & 183   & 56   & 72 & 21 & 32 & 740   \\
9 & 0.034  & 35 & 142  & 142   & 72   & 74 & 34 & 32 & 242   \\
10 & 0.0322& 45 & 9418 & 10001 & 1852 & 98 & 39 & 39 & 16435 \\
\hline
Avg.& 0.0547& 26.6 & 1415.6 & 1519.5 & 272.4 & 74.4 & 23.9 & 32.8 & 6557.44 \\
\hline 
\hline
\multicolumn{10}{c}{Decomposition Process Models} \\
\hline
1 & 0.0722 & 39  & 129   & 133   & 57    &  78 & 27 & 20 & 32000 \\
2 & 0.2325 & 120 &$ ^{*}$&$ ^{*}$&$ ^{*}$& 266 & 26 & 22 & 72270 \\
3 & 0.2167 & 156 &$ ^{*}$&$ ^{*}$&$ ^{*}$& 312 & 30 & 24 & 93240 \\
4 & 0.0745 & 40  & 862   & 900   & 282   & 93  & 26 & 24 & 79200 \\
5 & 0.088  & 80  &$ ^{*}$&$ ^{*}$&$ ^{*}$& 169 & 32 & 30 & 48600 \\
6 & 0.262  & 133 &$ ^{*}$&$ ^{*}$&$ ^{*}$& 283 & 27 & 20 & 23700 \\
7 & 0.0821 & 58  & 9501  & 10003 & 2140  & 128 & 26 & 30 & 65700 \\
8 & 0.5132 & 149 &$ ^{*}$&$ ^{*}$&$ ^{*}$& 349 & 17 & 20 & 31300 \\
9 & 0.0741 & 41  & 403   & 418   & 154   & 88  & 27 & 22 & 13455 \\
10 & 0.1705& 94  &$ ^{*}$&$ ^{*}$&$ ^{*}$& 180 & 24 & 22 & 89100 \\
\hline
Avg.& 0.1786& 91 & 2723.75 & 2863.5 & 658.25 & 194.6 & 26.2 & 23.4 & 54856.5 \\
\hline
\end{tabular}
\caption{Complexity Metrics for Our Process Models. $ ^{*}$: Reachability graph is unbounded. $ ^{**}$: Net is not a workflow net.}
\label{tab:metrics}
\end{table}

Now, given that for logs: 1, 2, 4 and 9, there is no preference of a particular group of models based on classification results; and given that Inductive Miner models are simpler than Decomposition models, we decided to consider Inductive Miner models for these logs, based on \emph{Occam's Razor} principle, 

In summary, our submitted models to the contest are: models 1, 2, 4, 8 and 9 of Inductive Miner models, and models 3, 5, 6, 7 and 10 of Decomposition models. The pictures of these process models are depicted in Appendix \ref{apx:models}


Based on this selection of process models, the contest organizers replayed the \emph{June} test logs. The reported classifications are shown in Table \ref{tab:JuneResults}, whereas the classification results are shown in Table \ref{tab:JuneResults2}. Our selection of models correctly classified 192 out of 200 traces. The organizers confirmed later that this submission gave us the \textbf{second} place in the contest among 14 participant teams.

\begin{table}[h]
\small
	\centering
		\begin{tabularx}{\textwidth}{L{1cm}|YY|YYYYYYYYYYYYYYYYYYYY}
 &  &  & \multicolumn{20}{c}{ Trace }\\
Log $\downarrow$& + & - & 1 & 2 & 3 & 4 & 5 & 6 & 7 & 8 & 9 & 10 & 11 & 12 & 13 & 14 & 15 & 16 & 17 & 18 & 19 & 20 \\
\hline
1 & 10 & 10 & $+$ & - & + & + & - & + & - & - & + & + & - & + & + & - & + & - & + & - & - & - \\
2 & 12 & 8  & - & - & + & + & - & \cellcolor{Gray}+ & + & + & - & \cellcolor{Gray}+ & + & - & + & + & - & - & + & + & + & - \\
3 & 9  & 11 & + & \cellcolor{Gray}- & + & - & - & - & + & + & + & + & + & - & - & - & - & - & + & - & + & - \\
4 & 10 & 10 & - & + & - & + & + & + & - & + & - & - & + & + & - & - & - & + & + & - & + & - \\
5 & 10 & 10 & - & + & - & - & - & - & - & - & + & - & + & + & + & - & + & + & + & + & - & + \\
6 & 12 & 8  & - & + & - & + & + & + & + & \cellcolor{Gray}+ & - & - & - & + & \cellcolor{Gray}+ & - & + & - & + & + & + & - \\
7 & 10 & 10 & + & + & - & - & - & + & - & + & + & + & - & + & - & - & - & - & + & + & + & - \\
8 & 13 & 7  & + & - & + & + & + & - & + & + & + & + & \cellcolor{Gray}+ & \cellcolor{Gray}+ & - & + & - & \cellcolor{Gray}+ & - & - & + & - \\
9 & 10 & 10 & + & + & + & + & + & - & - & + & - & + & - & + & - & - & - & + & - & - & - & + \\
10& 10 & 10 & + & + & - & - & - & + & - & + & + & + & + & - & + & + & - & - & + & - & - & - \\

	\end{tabularx}
	\caption{Reported Classifications of \emph{June} Logs. Shaded cells denote incorrectly classified traces.}
	\label{tab:JuneResults}
\end{table}


\begin{table}[h]
\small
	\centering
		\begin{tabularx}{\textwidth}{ L{3.4cm}|YYYYYYYYYY|Y}
 Log $\rightarrow$& 1 & 2 & 3 & 4 & 5 & 6 & 7 & 8 & 9 & 10 &  \\
		\hline
True Positive  & 10 & 10 & 9  & 10 & 10 & 10 & 10 & 10 & 10 & 10 & \\
False Positive & 0  & 2  & 0  & 0  & 0  & 2  & 0  & 3  & 0  & 0  & \\
True Negative  & 10 & 8  & 10 & 10 & 10 & 8  & 10 & 7  & 10 & 10 & \\
False Negative & 0  & 0  & 1  & 0  & 0  & 0  & 0  & 0  & 0  & 0  & \\
  \hline
Correctly Classified & 20 & 18 & 19 & 20 & 20 & 18 & 20 & 17 & 20 & 20 & \textbf{192}		\\	
		\end{tabularx}
	\caption{Classification Results of \emph{June} Logs.}
	\label{tab:JuneResults2}
\end{table}

\appendix

\section{Process Models}
\label{apx:models}

\begin{figure}[htb]
	\centering
		\includegraphics[width=1.00\textwidth]{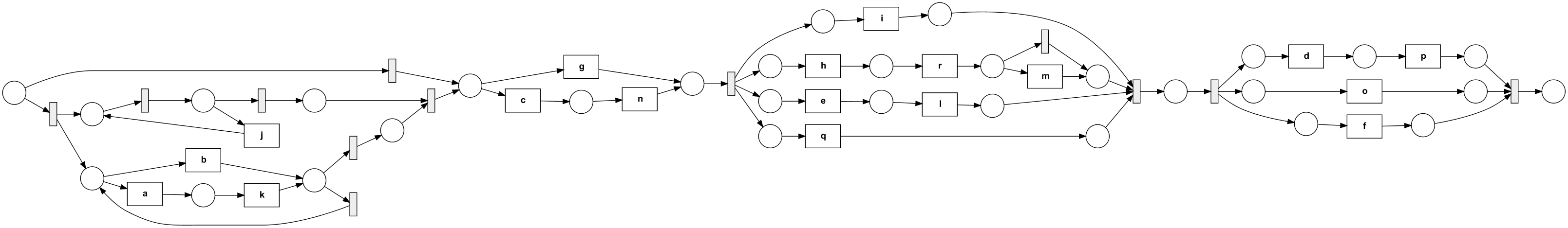}
	\caption{Process Model 1 (Inductive Miner)}
	\label{fig:m1}
\end{figure}

\begin{figure}[htb]
	\centering
		\includegraphics[width=1.00\textwidth]{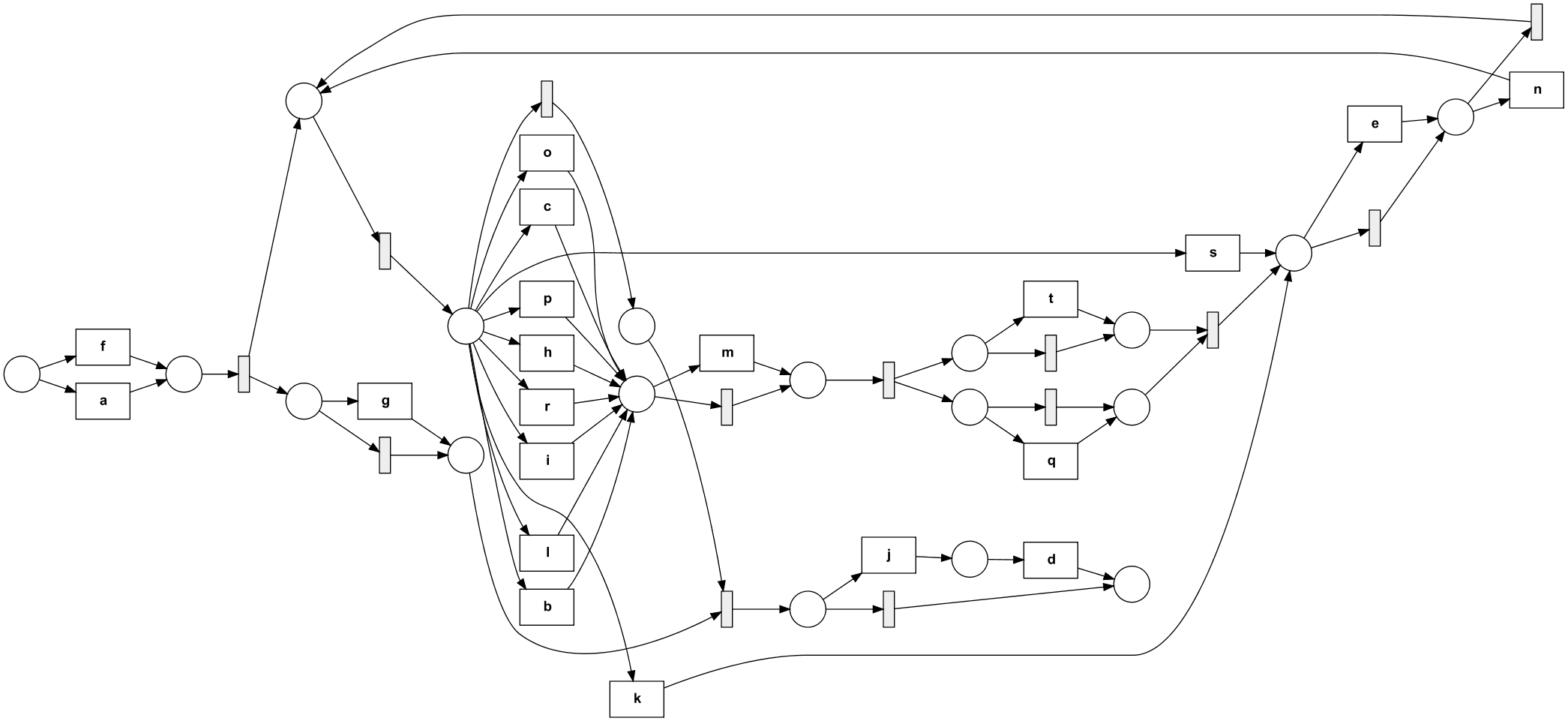}
	\caption{Process Model 2 (Inductive Miner)}
	\label{fig:m2}
\end{figure}

\begin{figure}[htb]
	\centering
		\includegraphics[width=1.00\textwidth]{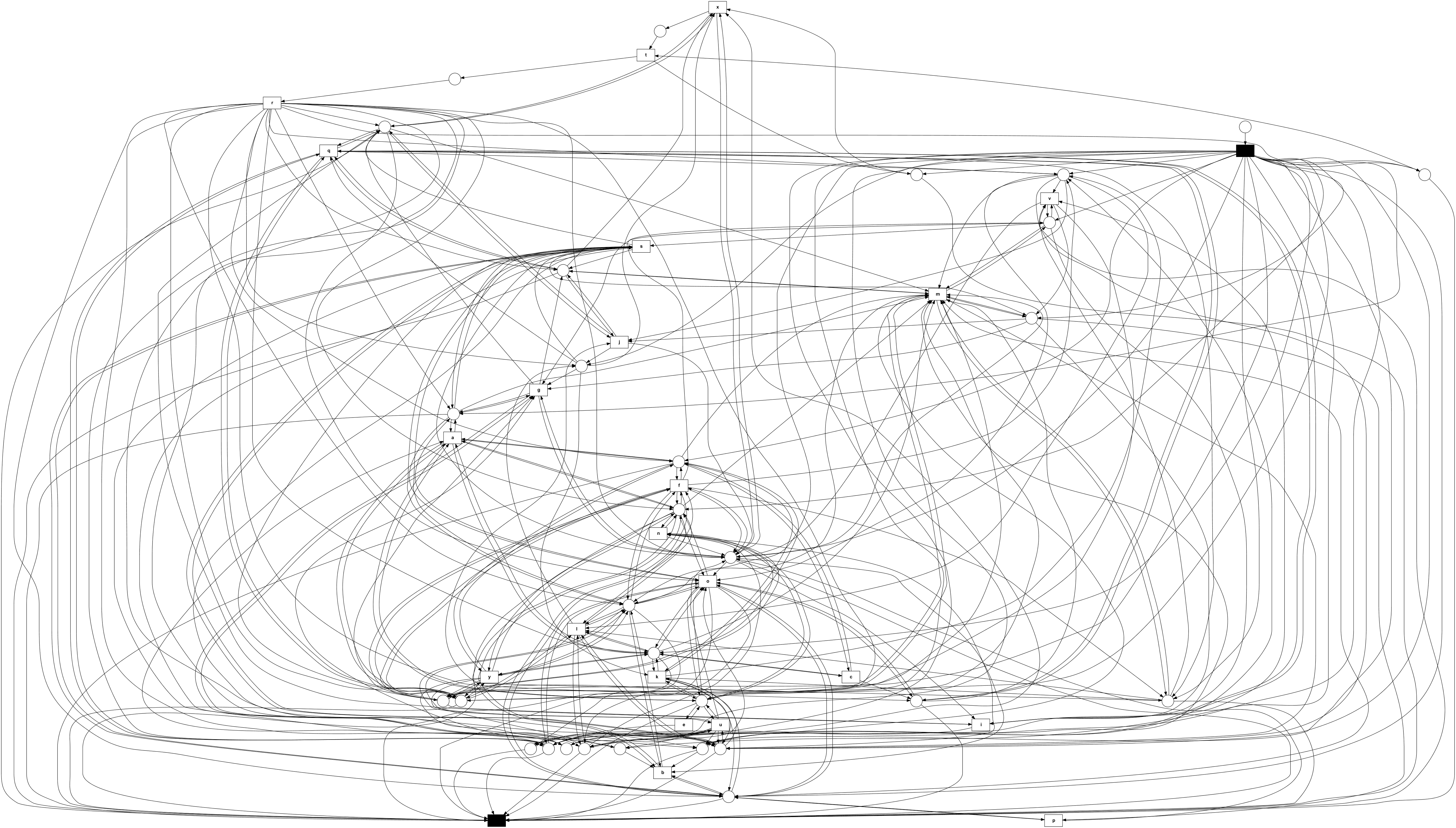}
	\caption{Process Model 3 (Decomposition)}
	\label{fig:m3}
\end{figure}

\begin{figure}[htb]
	\centering
		\includegraphics[width=1.00\textwidth]{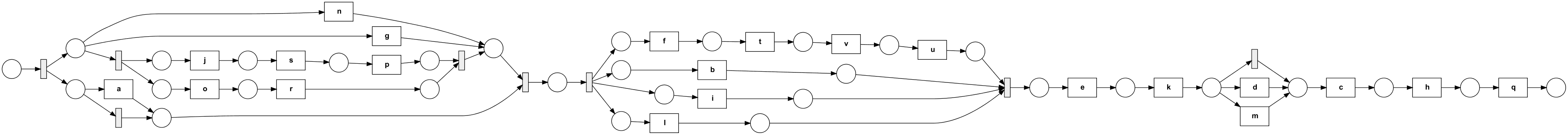}
	\caption{Process Model 4 (Inductive Miner)}
	\label{fig:m4}
\end{figure}

\begin{figure}[htb]
	\centering
		\includegraphics[width=1.00\textwidth]{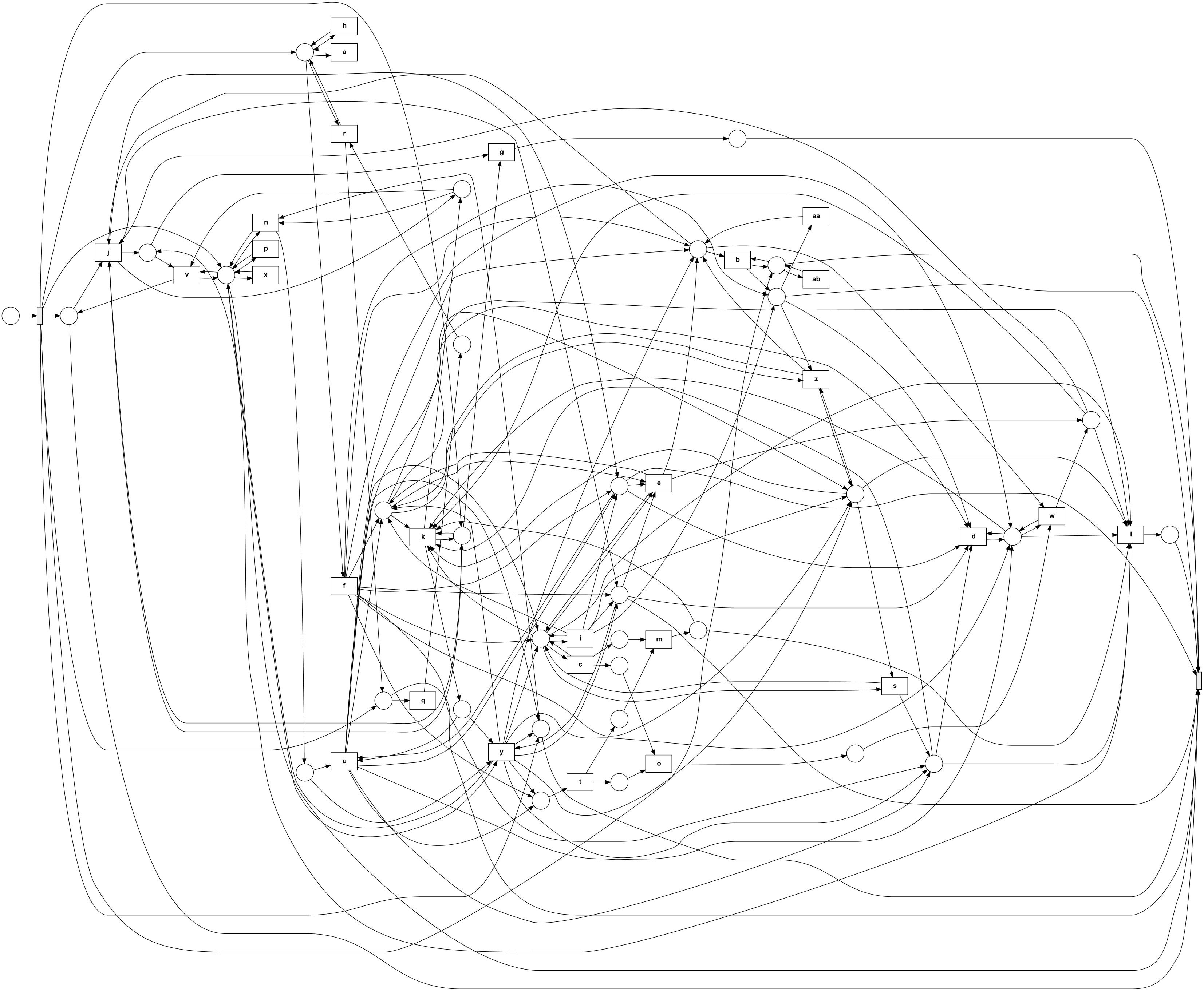}
	\caption{Process Model 5 (Decomposition)}
	\label{fig:m5}
\end{figure}

\begin{figure}[htb]
	\centering
		\includegraphics[width=1.00\textwidth]{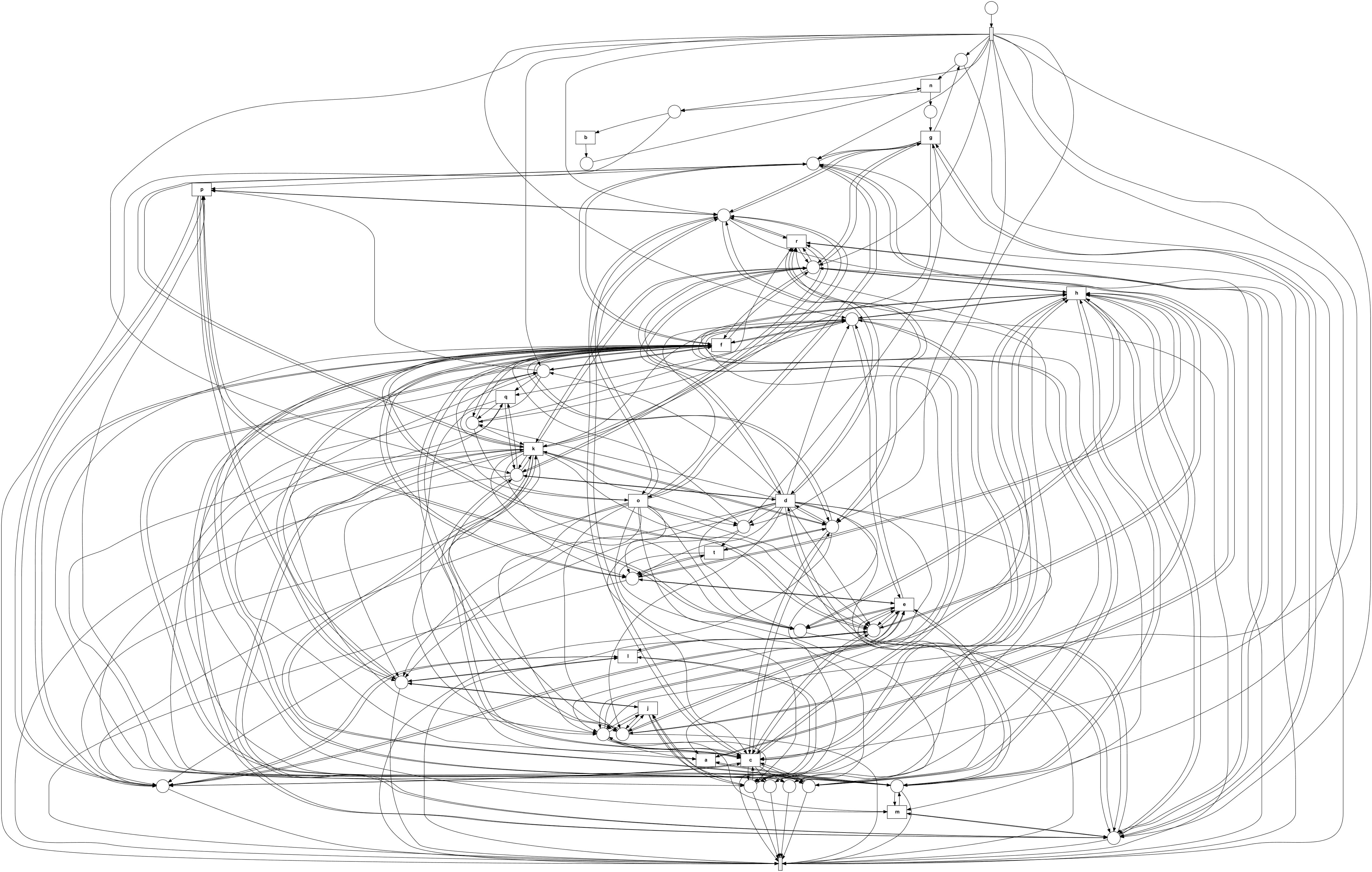}
	\caption{Process Model 6 (Decomposition)}
	\label{fig:m6}
\end{figure}

\begin{figure}[htb]
	\centering
		\includegraphics[width=1.00\textwidth]{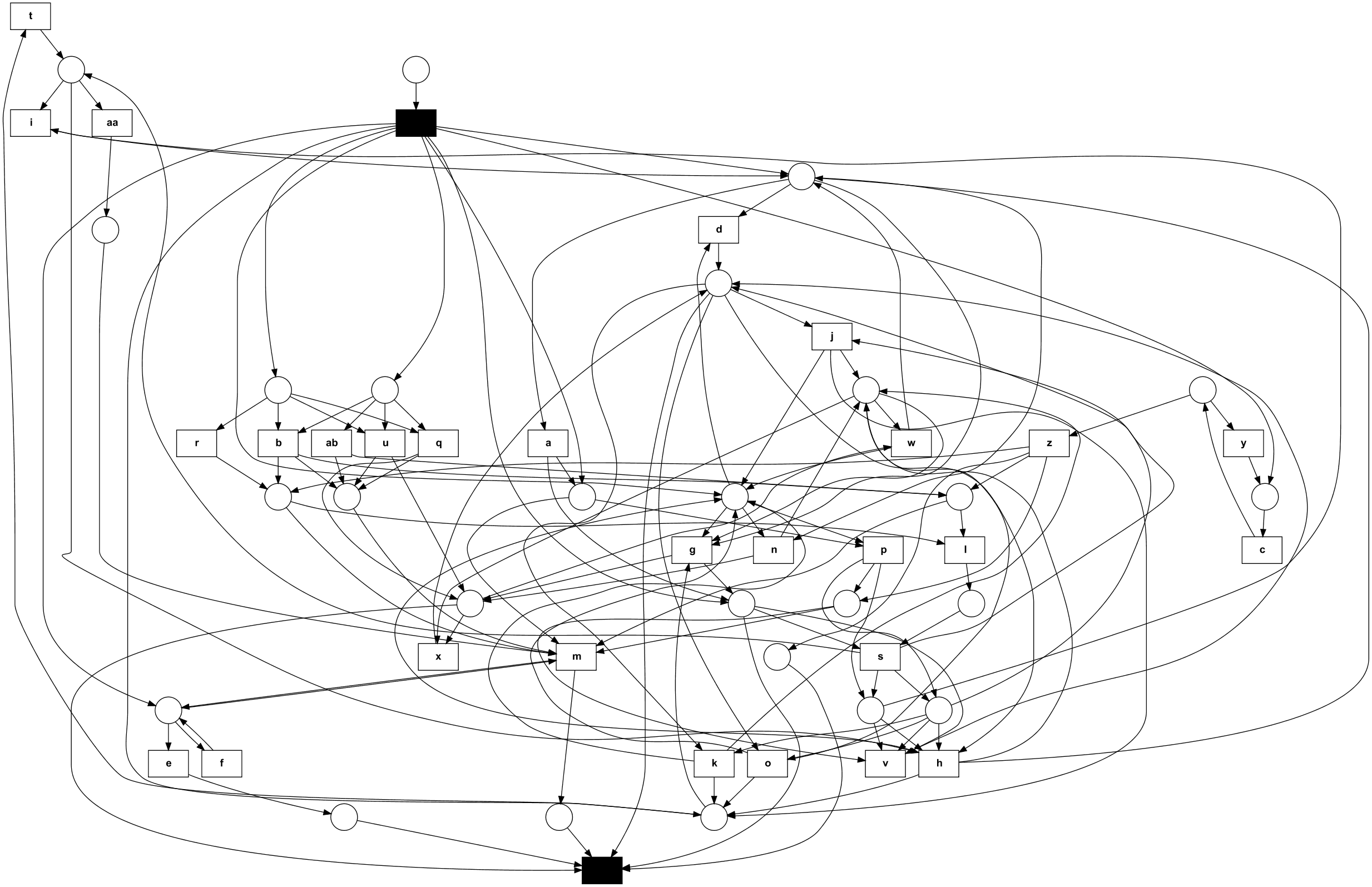}
	\caption{Process Model 7 (Decomposition)}
	\label{fig:m7}
\end{figure}

\begin{figure}[htb]
	\centering
		\includegraphics[width=1.00\textwidth]{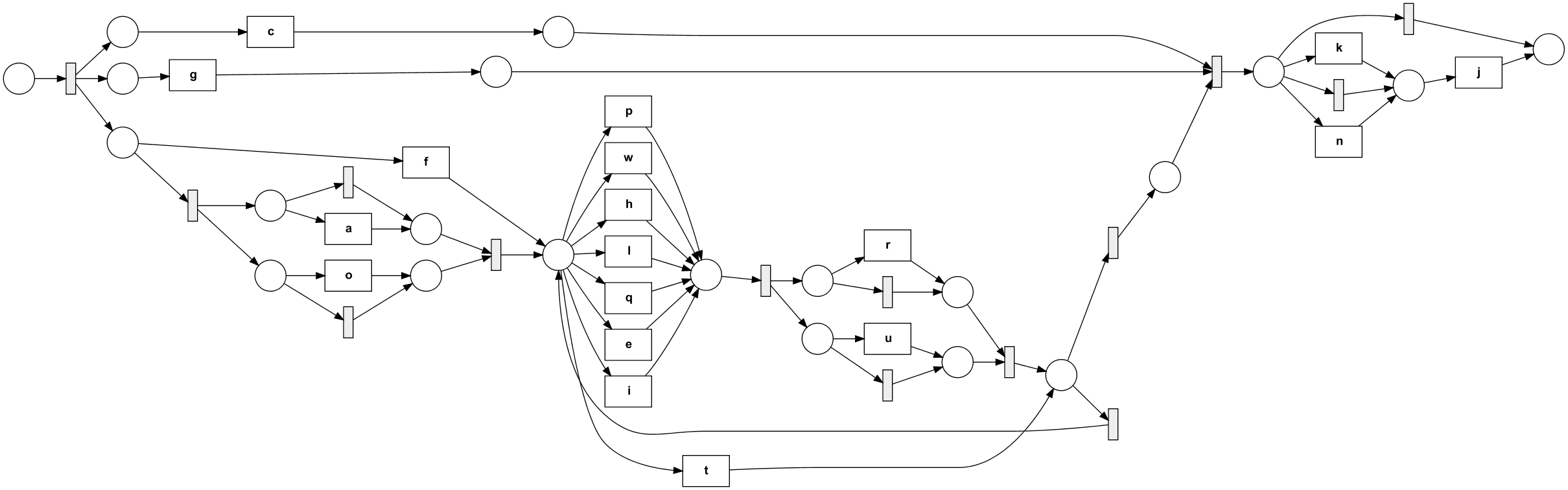}
	\caption{Process Model 8 (Inductive Miner)}
	\label{fig:m8}
\end{figure}

\begin{figure}[htb]
	\centering
		\includegraphics[width=1.00\textwidth]{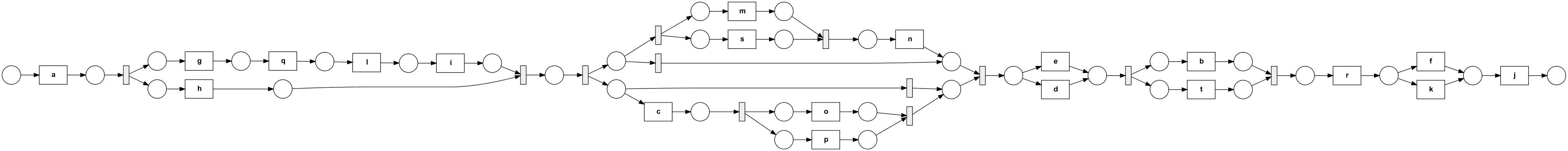}
	\caption{Process Model 9 (Inductive Miner)}
	\label{fig:m9}
\end{figure}

\begin{figure}[htb]
	\centering
		\includegraphics[width=1.00\textwidth]{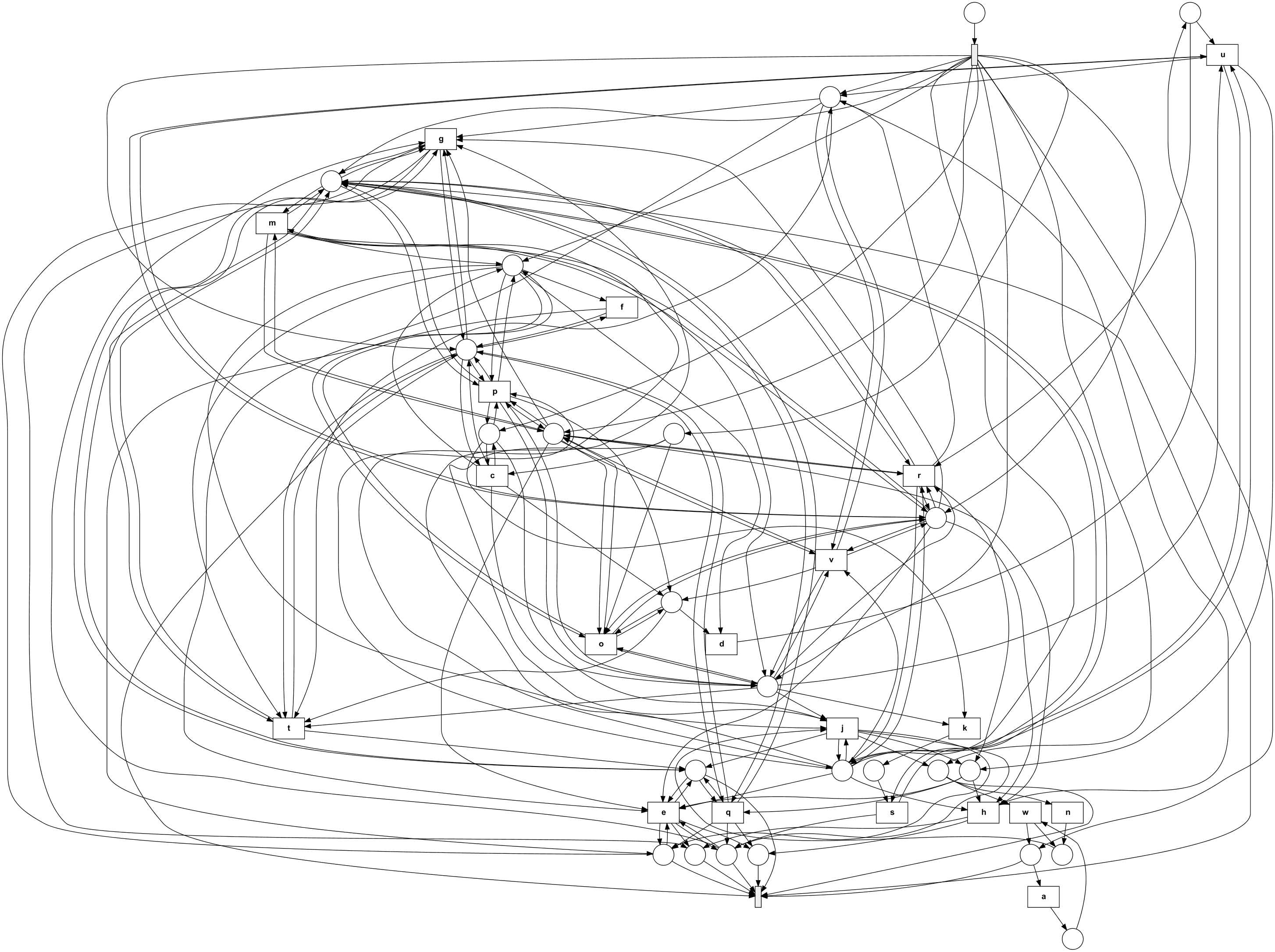}
	\caption{Process Model 10 (Decomposition)}
	\label{fig:m10}
\end{figure}

\end{document}